\documentclass{article}

\usepackage[preprint]{neurips_2026}

\usepackage[utf8]{inputenc} 
\usepackage[T1]{fontenc}    
\usepackage{hyperref}       
\usepackage{url}            
\usepackage{booktabs}       
\usepackage{amsfonts}       
\usepackage{nicefrac}       
\usepackage{microtype}      
\usepackage{xcolor}         
\usepackage{textcomp}
\usepackage{stfloats}
\usepackage{url}
\usepackage{verbatim}
\usepackage{graphicx}
\usepackage{caption}
\usepackage{booktabs}
\usepackage{tabularx}
\usepackage{multirow}
\usepackage{array}
\usepackage{amsmath}
\usepackage{xcolor}       
\usepackage{colortbl}     
\usepackage{array}
\usepackage{arydshln} 
\usepackage{breqn}
\usepackage{cuted}
\usepackage{float}
\usepackage[blocks]{authblk} 

\title{HeteroGenManip: Generalizable Manipulation For Heterogeneous Object Interactions }

\author{
     \textbf{Zhenhao Shen}$^{1\ast}$ \: \textbf{Zeming Yang}$^{2\ast}$ \: \textbf{Yue Chen}$^1$ \: \textbf{Yuran Wang}$^1$\\
    \vspace{-3mm}
    \textbf{Shengqiang Xu}$^1$ \: \textbf{Mingleyang Li}$^1$ \: \textbf{Hao Dong}$^1$ \: \textbf{Ruihai Wu}$^{1\dagger}$ 
    
    \normalsize $^1$Peking University \quad $^2$Tianjin University

    $^{\ast}$ Equal contribution \: ${^\dagger}$ Corresponding author \: Contact email: \href{zhshen25@stu.pku.edu.cn}{zhshen25@stu.pku.edu.cn}

    Website: \href{https://yzmyalier.github.io/HeteroGenManip/}{https://yzmyalier.github.io/HeteroGenManip/}
}

\date{} 

\begin{document}

\maketitle


\begin{abstract}
Generalizable manipulation involving cross-type object interactions is a critical yet challenging capability in robotics. To reliably accomplish such tasks, robots must address two fundamental challenges: ``where to manipulate'' (contact point localization) and ``how to manipulate'' (subsequent interaction trajectory planning). Existing foundation-model-based approaches often adopt end-to-end learning that obscures the distinction between these stages, exacerbating error accumulation in long-horizon tasks. Furthermore, they typically rely on a single uniform model, which fails to capture the diverse, category-specific features required for heterogeneous objects.
To overcome these limitations, we propose HeteroGenManip, a task-conditioned, two-stage framework designed to decouple initial grasp from complex interaction execution. First, Foundation-Correspondence-Guided Grasp module leverages structural priors to align the initial contact state, thereby significantly reducing the pose uncertainty of grasping. Subsequently, Multi-Foundation-Model Diffusion Policy (MFMDP) routes objects to category-specialized foundation models, integrating fine-grained geometric information with highly-variable part features via a dual-stream cross-attention mechanism. 
Experimental evaluations demonstrate that HeteroGenManip achieves robust intra-category shape and pose generalization. The framework achieves an average 31\% performance improvement in simulation tasks with broad type setting, alongside a 36.7\% gain across four real-world tasks with different interaction types.

\end{abstract}
\section{Introduction}

Real-world robotic applications often demand generalizable manipulation of heterogeneous objects, encompassing rigid, deformable, and articulated entities, and their complex cross-type interactions (\emph{e.g.}, hanging deformable garments on rigid clotheslines, inserting rigid hangers into soft fabrics). To reliably accomplish such tasks, robots must first address two fundamental questions that directly define the generalizable capability: where to manipulate (contact point localization) and how to manipulate (subsequent interaction trajectory planning). The difficulty of answering these questions stems from the need for robust perception and control across objects with drastically divergent physical properties (rigidity vs. deformability) and arbitrary shape variations, factoring that make different-type interaction manipulation far more complex than single-type scenarios.

\begin{figure}[t] 
  \centering
  \includegraphics[width=\columnwidth]{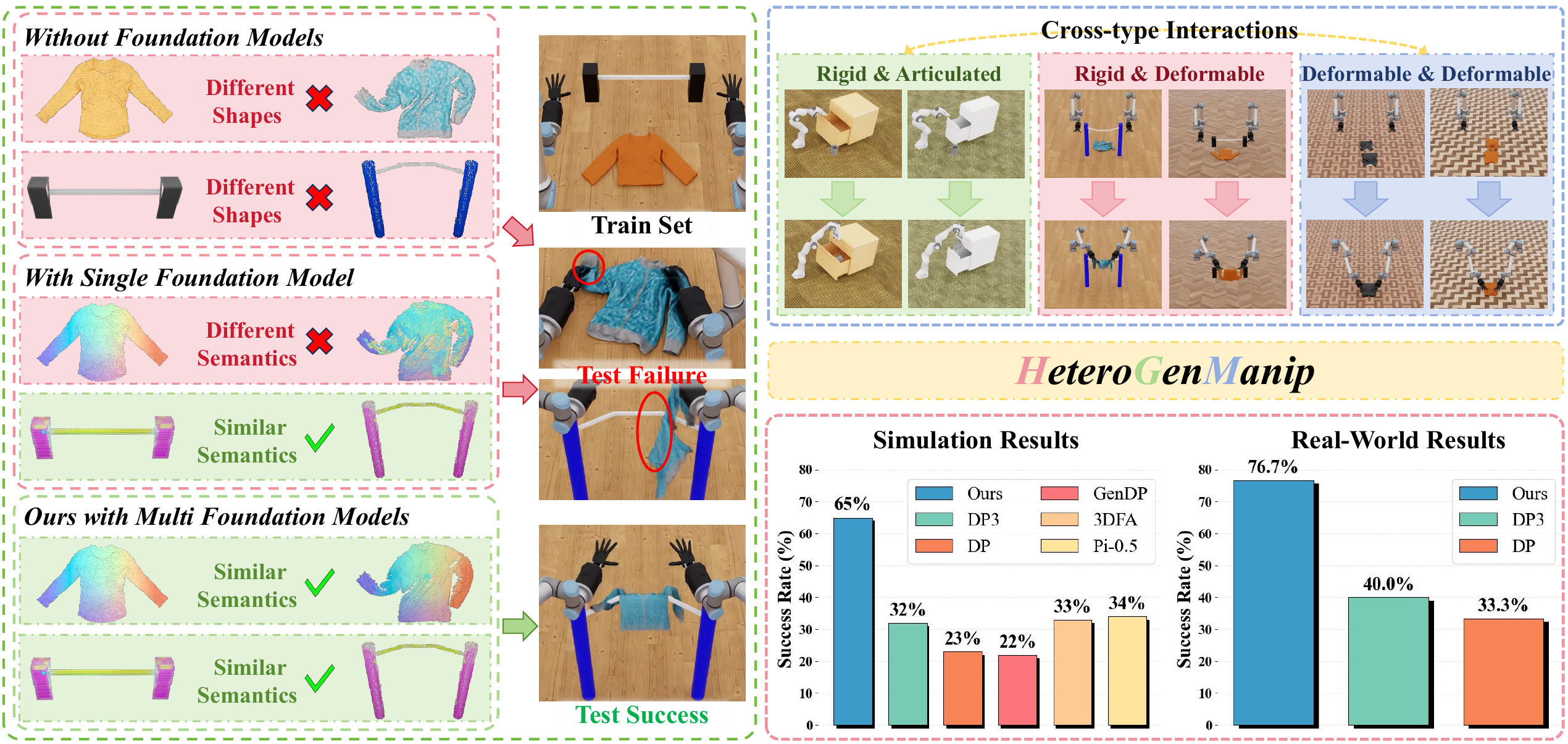} 
  \caption{
      \textbf{Motivation and Overview of HeteroGenManip.}
    \textbf{Left:} Comparison of three approaches for manipulation with different object types: 1) without foundation models, shape variations cause failure; 2) with a single foundation model, semantic understanding across object types is insufficient; 3) with multiple foundation models (ours), category-specific features enable successful manipulation.
    \textbf{Right:} HeteroGenManip handles different-type interactions and achieves state-of-the-art performance.
  }
  \label{fig:insight} 
  \vspace{-1em}
\end{figure}

To tackle this challenge, many existing methods~\cite{wang2023gendp, ju2025robo, gkanatsios20253dflowmatchactorunified, ke20243ddiffuseractorpolicy, Chen_2025_CVPR, wang2025dexgarmentlab, wang2025skil, miao2025knowledge} leverage the strong feature extraction and knowledge transfer capabilities of foundation models for generalization, showing promise in simple single-type manipulation. However, they hit a critical bottleneck when scaling to heterogeneous interactions, stemming from two core deficiencies:
first, they fail to decompose the manipulation process into specialized stages, instead, they adopt end-to-end learning that blurs the boundary between contact point localization and trajectory planning. This leads to redundant motion modeling, error accumulation in long-horizon tasks, and heightened learning difficulty.
Second, as shown in Fig.~\ref{fig:insight}, they usually rely on a single foundation model for all object types, overlooking the fact that different models excel at capturing category-specific features: rigid-focused models struggle with the non-rigid shape changes of deformable objects, while deformation-aware models are limited by the narrow range of deformable objects they target and fail to cover broad real-life objects. This mismatch results in inadequate representations for heterogeneous objects, limiting generalization across heterogeneous object interactions.

To address these problems, we propose \textbf{HeteroGenManip}, a task-conditioned, two-stage universal framework explicitly designed to resolve the ``where/how'' dual challenges of different-type interaction scenarios. Its core insight is to tailor foundation model usage to both the two manipulation stages and object type characteristics: using structural priors to decompose long-horizon tasks, and integrating category-specific models to match distinct object representation needs. These ideas are instantiated via two synergistic components:

1. \textit{Foundation-Correspondence-Guided Grasp}: this module addresses the ``where to manipulate'' question by leveraging structural priors to identify optimal contact points across heterogeneous objects. By decoupling localization from trajectory planning, it simplifies policy learning and reduces state-action space complexity. 
         
2. \textit{Multi-Foundation-Model Diffusion Policy (MFMDP)}: targeting the ``how to manipulate'' question, MFMDP employs category-specific foundation models to extract tailored features for different object types. This resolves the limitations of single-model approaches and enables effective modeling of cross-type interactions in trajectory planning.
        
We validate HeteroGenManip through extensive experiments on established benchmarks (\emph{e.g.}, RoboTwin~\cite{chen2025robotwin}, DexGarmentLab~\cite{wang2025dexgarmentlab}) and novel different-type interaction tasks, covering diverse object types and complex cross-type interactions. Results demonstrate significant improvements over state-of-the-art (SOTA) methods: it achieves an average 31\% performance improvement in simulation tasks with broad type setting, alongside a 36.7\% gain across four real-world tasks with different interaction types, confirming its robust generalization and real-world applicability.

In summary, our key contributions are fourfold:
\begin{itemize}
    \item We propose HeteroGenManip, a foundation model centered framework that addresses the generalization bottleneck in different-type interaction manipulation via two ideas: decomposing long-horizon tasks into two stages, and integrating multiple category-specific models.  
    \item We design Foundation-Correspondence-Guided Grasp that adapts to heterogeneous objects, leveraging structural priors to simplify workflows, reduce policy complexity, and explicitly answer the ``where to manipulate'' question.
    \item We develop the Multi-Foundation-Model Diffusion Policy (MFMDP), which embeds a category-specific feature extraction strategy and tailors foundation models to object type needs, addressing the ``how to manipulate'' challenge and resolving single model limitations.
    \item We validate HeteroGenManip on a comprehensive test suite, demonstrating superior generalization over SOTA methods (31\% simulation gain, 36.7\% real-world gain) and providing a new paradigm for heterogeneous object interaction manipulation.
\end{itemize}


\section{Related Work}
\label{rw}
\subsection{Multi-Object Interaction Manipulation}
Recent research on Multi-Object Interaction Manipulation has focused on two directions based on object physical properties: works like \cite{chen2025robotwin, Mu_2025_CVPR, james2019rlbenchrobotlearningbenchmark, grotz2024peract2benchmarkinglearningrobotic, zeng2020transporter, li2024behavior1k, gu2023maniskill2, robosuite2020, mclean2025metaworld, gymnasium_robotics2023github, mo2021o2oafford, yuan2024learningmanipulateanywherevisual} concentrate on rigid-rigid interaction, developing benchmark or general methods for tasks like rigid component placement; while studies including \cite{wang2025dexgarmentlab, wu2025garmentpile, lu2024garmentlab, li2023dexdeformdexterousdeformableobject, huang2021plasticinelab} focus on rigid-deformable interaction, addressing deformation-aware manipulation of deformable objects. However, a limitation exists: most approaches rely on global scene features or individual object features for policy learning, failing to explicitly leverage the interaction features between the two manipulated objects. This oversight undermines performance in tasks requiring precise inter-object coordination. Thus, our work aims to propose a general framework suitable for manipulation with heterogeneous object interactions.

\subsection{Imitation Learning with Foundation Model}
Policies relying on pure image or point cloud inputs have achieved promising results in simple robotic manipulation scenarios~\cite{chi2024diffusionpolicy, Ze2024DP3, zhao2023learningfinegrainedbimanualmanipulation,tie2025etseed, goyal2024rvt2learningprecisemanipulation, zhang2024leveraging, wang2024rise, lu2025h3dptriplyhierarchicaldiffusionpolicy, huang2025prismpointcloudreintegratedinference, lu2024manicm, prasad2024consistency}. However, due to insufficient data and limited model scales, they lack generalization capabilities. Recent studies have made significant progress in enhancing policy generalization using foundation models. GenDP~\cite{wang2023gendp}, G3Flow~\cite{Chen_2025_CVPR} and Know~\cite{miao2025knowledge} leverage DINOv2~\cite{oquab2023dinov2} for feature extraction. SKIL~\cite{wang2025skil} employs DIFT~\cite{tang2023emergent} with SAM~\cite{kirillov2023segany} for key point extraction. 3DDA~\cite{ke20243ddiffuseractorpolicy} and 3DFA~\cite{gkanatsios20253dflowmatchactorunified} extract CLIP~\cite{radford2021learningtransferablevisualmodels} features from 2D images and lift them to 3D using Act3D's methodology~\cite{gervet2023act3d3dfeaturefield}. Nevertheless, these works mainly focus on rigid objects, while HALO~\cite{wang2025dexgarmentlab} utilizes GAM for deformable objects. However, all these methods rely on a single foundation model for all object types, overlooking that different models excel at capturing category-specific features. Therefore, we propose a robot manipulation policy based on multiple foundation models to realize generalization.

\subsection{Manipulation with Correspondence}
Recent studies leverage corresponding points to guide robotic manipulation. Existing methods fall into two categories: one pursues zero-shot acquisition via diffusion models~\cite{zhang2023talefeaturesstablediffusion, ye2023affordance, Dutt_2024_CVPR, sharp2022diffusionnetdiscretizationagnosticlearning,luo2024diffusionhyperfeaturessearchingtime,
tang2023emergentcorrespondenceimagediffusion} or encoder-decoder architectures (e.g., Where2Act~\cite{mo2021where2actpixelsactionsarticulated} with UNet~\cite{ronneberger2015unetconvolutionalnetworksbiomedical} and PointNet++~\cite{qi2017pointnetdeephierarchicalfeature}); the other infers correspondences via cosine similarity with prior objects~\cite{wang2025dexgarmentlab, ju2025robo, wu2025garmentpile, Wu_2024_CVPR, wu2024afforddp}. The former enables zero-shot capability but suffers from complex architectures and low efficiency; the latter offers a concise framework. Otherwise, most works focus on rigid objects~\cite{ju2025robo, geng2023rlafford, ye2023affordance, hou2021affordance, zhu2024densematcherlearning3dsemantic, florence2018denseobjectnetslearning, kuang2024ram, xue2023useekunsupervisedse3equivariant3d}, with few extending to deformable objects~\cite{wang2025dexgarmentlab, wu2025garmentpile, Wu_2024_CVPR, Chen_2023}. Our work achieves generalized grasp across both rigid and deformable objects via an efficient corresponding-point-based approach.

\begin{figure*}[ht] 
  \centering
  \includegraphics[width=\textwidth]{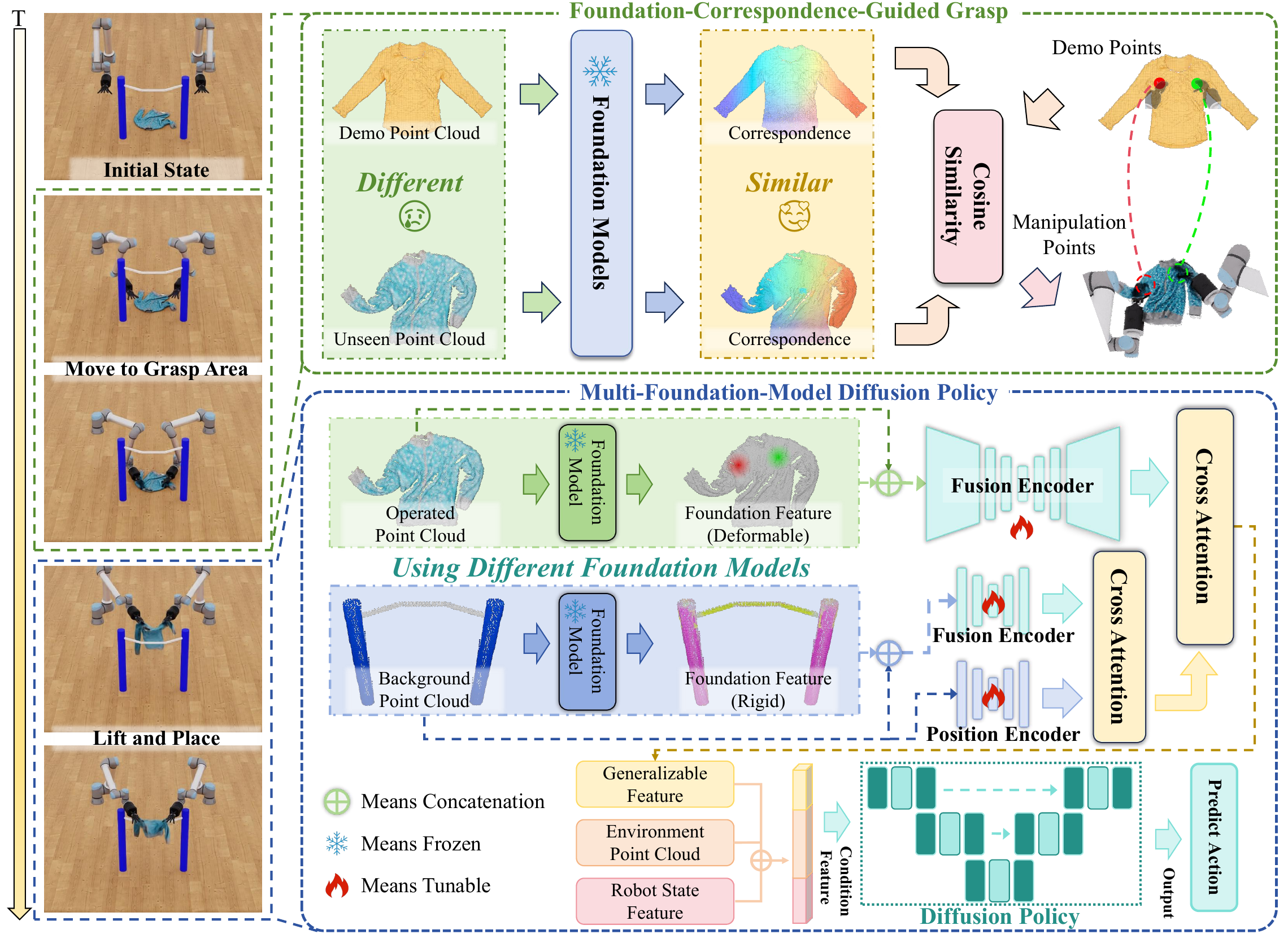} 
  \caption{
  \textbf{HeteroGenManip Architecture.} Our framework comprises two phases: Foundation-Correspondence-Guided Grasp and Multi-Foundation-Model Diffusion Policy. First, we leverage foundation model correspondence to identify manipulation points and execute grasping. Next, we select category-specific foundation models for feature extraction based on object types, then integrate the features via the Fusion Module into the Diffusion Policy for action prediction.
  }
  \label{fig:arch} 
  \vspace{-1em}
\end{figure*}

\section{Methods}
\label{method}
In this section, we detail the different components of our HeteroGenManip approach, as shown in Fig.~\ref{fig:arch}.
We elaborate on the module for conducting precise grasp for target objects in the Foundation Correspondence Guided Grasp (Section~\ref{CG}). We then introduce the policy that generates complex post-grasp trajectories in the Multi-Foundation-Model Diffusion Policy (Section~\ref{MFMDP}). For foundation models, we use Garment Affordance Model (GAM)~\cite{wang2025dexgarmentlab} and Uni3D~\cite{zhou2023uni3d}, whose detailed descriptions and selection criteria are provided in Appendix~\ref{appendix:foundation_models}.

\subsection{Foundation-Correspondence-Guided Grasp}
\label{CG}
To improve the efficiency and robustness of imitation learning, we leverage pre-trained foundation models to extract and match semantically consistent grasp regions. This spatial anchoring avoids modeling long, complex trajectories from scratch and effectively handles intra-category shape variations. Below, we detail this mechanism in two parts: localizing the manipulation point (Section~\ref{Corresponding Points}) and determining its grasping orientation (Section~\ref{Orientation-Aware Grasping}).

\subsubsection{Corresponding Points}
\label{Corresponding Points}
To determine the specific manipulation point on a target object, we rely on a pre-trained, category-specific foundation model $\mathcal{M}_{\mathcal{C}}$ corresponding to the object's category $\mathcal{C}$. This model extracts point-wise structural and semantic features from both the demonstration and target point clouds. 

Given a pre-defined manipulation point in the demonstration, we identify its exact counterpart on the target object by computing the cosine similarity between their extracted foundation features. The point in the target point cloud that maximizes this similarity is designated as the target manipulation point. This feature-level matching ensures that category-specific semantic information (\emph{e.g.}, a cup handle) is robustly captured and localized, regardless of the target object's geometric variations.

\subsubsection{Orientation-Aware Grasping}
\label{Orientation-Aware Grasping}
After localizing the manipulation point, determining the correct grasp orientation is crucial. Unlike prior methods (\emph{e.g.}, Robo-ABC~\cite{ju2025robo}) that rely on external grasp planners (\emph{e.g.}, AnyGrasp~\cite{fang2023anygrasprobustefficientgrasp}), we integrate orientation estimation directly into our framework to maximize efficiency.

For deformable objects ($\mathcal{O}_{\text{deform}}$) in flat states, a fixed orientation vector suffices. For rigid objects ($\mathcal{O}_{\text{rigid}}$) with rotational variations, we introduce auxiliary reference points during demonstration. Matching these points on the target object constructs a directional vector that determines the 3D rotational orientation, eliminating the need for additional pose-estimation modules. This approach provides a highly reliable grasp foundation for both grippers and dexterous hands.

\subsection{Multi-Foundation-Model Diffusion Policy}
\label{MFMDP}
Following the initial spatial anchoring, the Multi-Foundation-Model Diffusion Policy (MFMDP) subsequently generates complex interaction trajectories. To effectively model cross-type interactions, MFMDP must reconcile the heterogeneous semantic features extracted by different category-specific foundation models, and fuse them with dynamic spatial information. Below, we detail this mechanism in three parts: establishing unified point descriptors (Section~\ref{point desc}), fusing features via dual-stream cross-attention (Section~\ref{fusion module}), and integrating holistic contextual streams (Section~\ref{holistic integration}).

\subsubsection{3D Point Description}
\label{point desc}
Different foundation models yield distinct feature representations. To establish a unified point descriptor across object types, we apply tailored processing strategies before integrating them with spatial coordinates. For deformable objects ($\mathcal{O}_{\text{deform}}$), we leverage the Garment Affordance Model (GAM)~\cite{wang2025dexgarmentlab}. Instead of using raw high-dimensional outputs, we compute the cosine similarity between each point's GAM feature and a set of predefined demonstration points. This inherently yields a compact, similarity-based feature vector that captures topological relations. 
For rigid objects ($\mathcal{O}_{\text{rigid}}$), foundation models like Uni3D~\cite{zhou2023uni3d} output exceptionally high-dimensional semantic vectors. Directly concatenating these with 3D coordinates would cause a severe dimensional imbalance, where dominant semantic features overshadow critical fine-grained positional cues. To resolve this, we employ PCA to compress the Uni3D features into a low-dimensional space. 

Finally, for both object types, we concatenate these processed, low-dimensional semantic features with their corresponding raw 3D coordinates ($\mathbf{p} \in \mathbb{R}^3$). This forms a unified point descriptor that balances high-level category semantics with exact geometric positions.

\subsubsection{Cross-Attention Based Fusion Module}
\label{fusion module}
A critical limitation in conventional representation learning is that encoding point descriptors leads to
semantic features dominating the latent space, impairing spatial geometric learning. To address this, we propose a dual-stream encoding mechanism to explicitly reinforce spatial understanding.
  
We split the target object's point cloud into two streams: a \textit{Spatial Coordinate Stream} (encoding raw 3D coordinates to preserve geometry) and a \textit{Semantic Descriptor Stream} (encoding the unified descriptors). We fuse these streams using cross-attention, where semantic descriptors serve as the query, and spatial coordinates act as the key and value. This ensures semantic attributes enhance, rather than suppress, spatial geometric cues.

We optimize this architecture based on our two-stage pipeline. Since the operated object (held by the gripper) has been explicitly localized during grasping, it does not require redundant spatial reinforcement; we directly encode its semantic descriptors. The background object (\emph{e.g.}, clothesline or plate), however, lacks this prior positioning and passes through the full dual-stream encoder. To capture interaction dynamics (\emph{e.g.}, supporting or inserting), we perform inter-object cross-attention fusion, using the operated object's features as query and the background object's features as key and value, yielding a comprehensive relational feature map.

\subsubsection{Holistic Feature Integration}
\label{holistic integration}
For the final state representation, we integrate three contextual streams. Following DP3~\cite{Ze2024DP3}, we encode the global dynamic point cloud and robot joint states via MLP. We then concatenate these with the relational feature from the Cross-Attention Fusion Module.

This holistic feature vector unifies dynamic environmental context, robot states, and category-specific object semantics. It is then fed into a DDIM-based diffusion model to iteratively denoise and predict the optimal subsequent interaction actions. Training details can be found in
Appendix~\ref{appendix:training}.

\section{Simulation Experiments}
\label{simulation}
\begin{figure*}[t] 
  \centering
  \includegraphics[width=\linewidth]{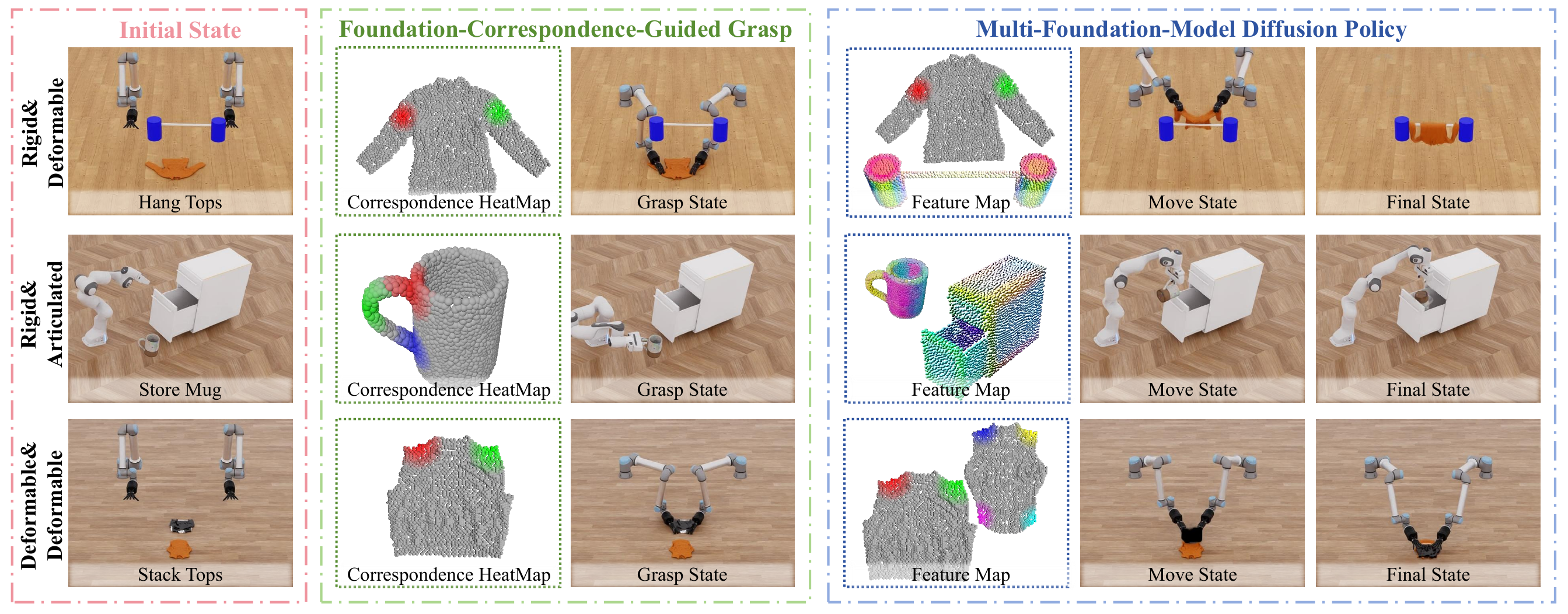} 
  \caption{\textbf{Whole Procedure Of Our Framework.} Three representative tasks with distinct interaction types demonstrate the full workflow of our policy, which consists of four states: Initial State, Grasp State, Move State, and Final State. The detailed execution description is in Appendix~\ref{appendix:frameworkexec}.}
  \label{fig:whole_procedure} 
  \vspace{-0.5em}
\end{figure*}
\subsection{Setup}

\begin{table*}[t]
  \centering
  \caption{Simulation Results (Mean±Std over 3 runs). \textbf{Bold} indicates the best performance, and \underline{underline} indicates the second-best.}
  \label{tab:simulation_result}
  \small
  \renewcommand{\arraystretch}{1.2}
  \definecolor{task2}{RGB}{240,240,240}

  \begin{tabular*}{\textwidth}{@{\extracolsep{\fill}} l l *{6}{c}}

    \toprule
    Task / Method & Setting &  DP & DP3 & GenDP-S & 3DFA & Pi0.5 & Ours \\
    \midrule

    \multirow{2}{*}{Stack Tops}
    & Train & \underline{0.39±0.05} & 0.37±0.05 & 0.32±0.06 & 0.35±0.05 & 0.29±0.04 & \textbf{0.75±0.01} \\
    & \cellcolor{task2}Test  & \underline{0.25±0.03} & 0.25±0.04 & 0.20±0.04 & 0.16±0.02 & 0.25±0.03 & \textbf{0.60±0.03} \\

    \multirow{2}{*}{Store Tops}
    & Train & \underline{0.29±0.05} & 0.21±0.05 & 0.13±0.05 & 0.17±0.04 & 0.17±0.04 & \textbf{0.57±0.03} \\
    & \cellcolor{task2}Test  & \underline{0.10±0.02} & 0.06±0.02 & 0.04±0.02 & 0.08±0.02 & 0.08±0.02 & \textbf{0.50±0.02} \\

    \multirow{2}{*}{Hang Tops}
    & Train & 0.27±0.05 & \underline{0.52±0.06} & 0.36±0.06 & 0.47±0.05 & 0.39±0.05 & \textbf{0.71±0.03} \\
    & \cellcolor{task2}Test  & 0.22±0.02 & 0.25±0.04 & 0.21±0.03 & \underline{0.27±0.04} & 0.24±0.04 & \textbf{0.62±0.04} \\

    \multirow{2}{*}{Hang Coats}
    & Train & 0.64±0.06 & \underline{0.83±0.05} & 0.58±0.06 & 0.73±0.05 & 0.45±0.05 & \textbf{0.91±0.03} \\
    & \cellcolor{task2}Test  & 0.39±0.03 & 0.57±0.04 & 0.34±0.04 & \underline{0.57±0.03} & 0.34±0.04 & \textbf{0.70±0.03} \\

    \multirow{2}{*}{Wear Bowl Hat}
    & Train & 0.50±0.06 & \underline{0.73±0.05} & 0.60±0.06 & 0.70±0.06 & 0.57±0.05 & \textbf{0.83±0.03} \\
    & \cellcolor{task2}Test  & 0.19±0.03 & 0.46±0.04 & 0.32±0.04 & \underline{0.50±0.04} & 0.46±0.04 & \textbf{0.68±0.02} \\

    \multirow{2}{*}{Store Mug}
    & Train & 0.35±0.05 & 0.45±0.05 & 0.23±0.05 & 0.43±0.05 & \underline{0.51±0.05} & \textbf{0.89±0.05} \\
    & \cellcolor{task2}Test  & 0.11±0.03 & 0.10±0.02 & 0.05±0.02 & 0.24±0.04 & \underline{0.32±0.04} & \textbf{0.64±0.03} \\

    \multirow{2}{*}{Place Bread}
    & Train & 0.48±0.06 & 0.83±0.05 & 0.50±0.06 & 0.73±0.05 & \underline{0.87±0.05} & \textbf{0.98±0.02} \\
    & \cellcolor{task2}Test  & 0.31±0.03 & 0.73±0.05 & 0.42±0.04 & 0.66±0.04 & \underline{0.76±0.04} & \textbf{0.98±0.02} \\

    \multirow{2}{*}{Beat Block (C)}
    & Train & 0.35±0.05 & 0.23±0.05 & 0.25±0.05 & 0.31±0.05 & \underline{0.39±0.05} & \textbf{0.65±0.03} \\
    & \cellcolor{task2}Test  & 0.27±0.03 & 0.13±0.03 & 0.16±0.04 & 0.17±0.03 & \underline{0.30±0.04} & \textbf{0.50±0.03} \\

    \midrule

    \multirow{2}{*}{Average}
    & Train & 0.41±0.05 & \underline{0.52±0.05} & 0.37±0.06 & 0.49±0.05 & 0.45±0.05 & \textbf{0.78±0.03} \\
    & \cellcolor{task2}Test  & 0.23±0.03 & 0.32±0.04 & 0.22±0.03 & 0.33±0.03 & \underline{0.34±0.04} & \textbf{0.65±0.02} \\

    \bottomrule
  \end{tabular*}
  \vspace{-1em}
\end{table*}

\subsubsection{Task Description}
We select and adapt partial tasks from DexGarmentLab \cite{wang2025dexgarmentlab} and RoboTwin \cite{chen2025robotwin}, and also develop several novel tasks based on DexGarmentLab. These tasks comprehensively cover different-type interactions, enabling a holistic evaluation of our model's capabilities. The detailed descriptions and visualization of these tasks are in Appendix~\ref{appendix:sim_tasks}.

\subsubsection{Baselines}
We compare our framework against five state-of-the-art baselines: 1) \textbf{DP}~\cite{chi2024diffusionpolicy}, a foundational RGB-conditioned diffusion policy; 2) \textbf{DP3}~\cite{Ze2024DP3}, which extends DP by utilizing compact 3D point cloud representations; 3) \textbf{GenDP-S}~\cite{wang2023gendp}, a single-camera variant of GenDP that predicts actions via diffusion over 3D semantic descriptor fields; 4) \textbf{3DFA}~\cite{gkanatsios20253dflowmatchactorunified}, a flow-matching framework utilizing 3D pre-trained visual representations; and 5) \textbf{Pi0.5}~\cite{intelligence2025pi05visionlanguageactionmodelopenworld}, a representative Vision-Language-Action (VLA) model based on flow matching.
Training details are provided in Appendix~\ref{appendix:baselines}.

\subsubsection{Data Collection and Evaluation}
For policy training, we collect 50 expert demonstrations for each task. During evaluation, we conduct 3 independent experimental runs, each performing 50 rollout attempts under randomly sampled initialization seeds. To ensure rigorous and fair comparison, all baseline methods are evaluated using identical environmental seed across the same run. We report the mean success rate and standard deviation across the 3 runs.

  
\begin{table*}[t]
  \centering
  \caption{Ablation Results (Mean±Std over 3 runs, Test setting only, D means Dinov2).}
  \label{tab:ablation}
  \renewcommand{\arraystretch}{1.2}

  \resizebox{\textwidth}{!}{
  \begin{tabular}{l *{9}{c}}
    \toprule
      Method / Task & \shortstack{Stack\\Tops} & \shortstack{Store\\Tops} & \shortstack{Hang\\Tops} & \shortstack{Hang\\Coats} & \shortstack{Wear\\Bowl Hat} & \shortstack{Store\\Mug} & \shortstack{Place\\Bread} & \shortstack{Beat\\Block(C)} & Average \\
    \midrule
    Ours w/o CG & 0.32±0.04 & 0.09±0.03 & 0.30±0.05 & 0.62±0.02 & 0.66±0.04 & 0.38±0.06 & 0.64±0.03 & 0.30±0.04 & 0.41±0.04 \\
    Ours w/o PE & 0.30±0.02 & 0.40±0.03 & 0.32±0.05 & 0.62±0.03 & 0.38±0.05 & 0.34±0.07 & 0.94±0.04 & 0.13±0.03 & 0.43±0.04 \\
    Ours w/o MFM & -- & -- & 0.30±0.05 & 0.32±0.02 & 0.34±0.04 & -- & -- & -- & -- \\
    Ours w/o MFM (D) & -- & -- & 0.24±0.04 & 0.15±0.03 & 0.20±0.02 & -- & -- & -- & -- \\
    Ours & \textbf{0.60±0.03} & \textbf{0.50±0.02} & \textbf{0.62±0.04} & \textbf{0.70±0.03} & \textbf{0.68±0.02} & \textbf{0.64±0.03} & \textbf{0.98±0.02} & \textbf{0.50±0.03} & \textbf{0.65±0.02} \\
    \bottomrule
  \end{tabular}
  }
  \vspace{-0.5em}
\end{table*}

\begin{figure*}[t] 
  \centering
  \includegraphics[width=\textwidth]{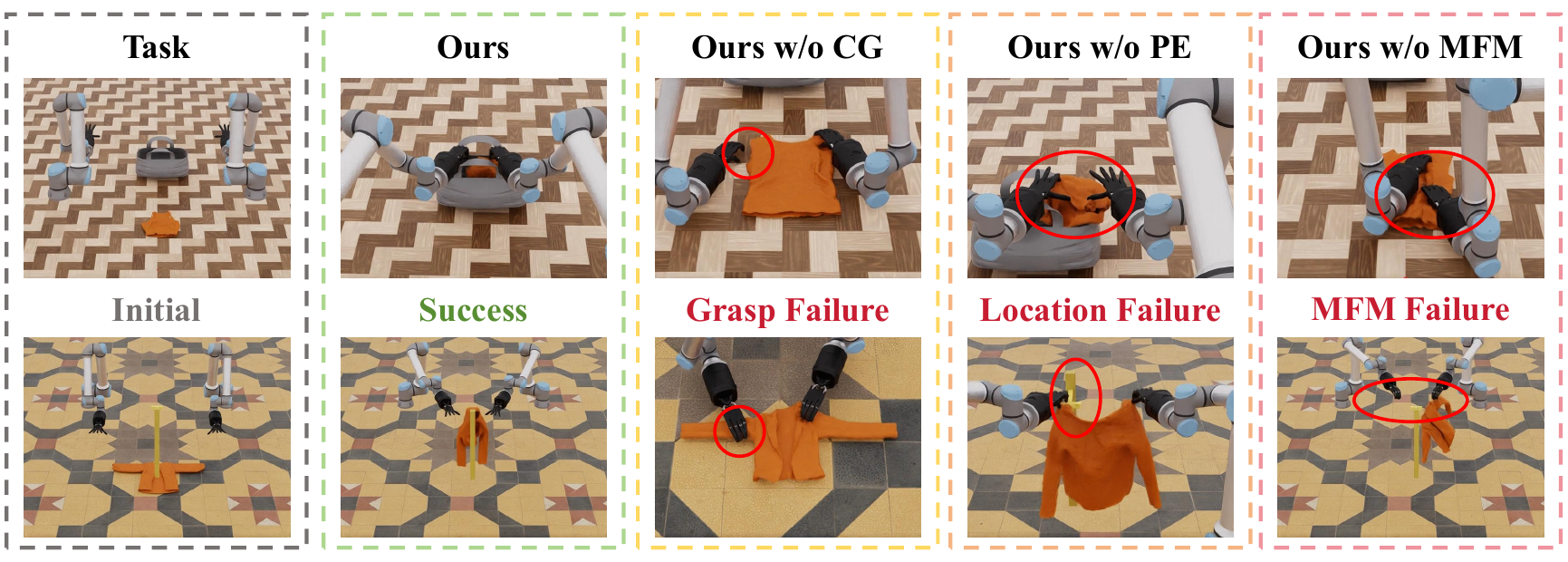} 
  \caption{\textbf{Ablation studies.} When foundation-correspondence-guided grasp is removed (w/o CG), the model fails to perform grasping during the grasp phase, making subsequent task completion impossible. Without the position enhancement encoder (w/o PE), the loss of positional information of background objects leads to positional deviations during the positioning and placement stage. When the multi-foundation-model architecture is excluded (w/o MFM), failures may occur at various stages of the entire pipeline, defined as MFM failure.}
  \label{fig:Ablation} 
  \vspace{-1em}
\end{figure*}

\subsection{Result and Analysis}

This section comprehensively evaluates the proposed framework (denoted as “Ours”) across multiple robot manipulation tasks under two experimental settings: Train (experiments conducted within the training distribution, featuring minor spatial variations and nearly fixed object shapes) and Test (experiments in novel environments, with larger spatial displacements and varying object shapes).

As shown in Table~\ref{tab:simulation_result}, our model achieves an average success rate of 0.78 in training scenarios, significantly outperforming the best baseline (0.52). More importantly, in unseen test environments, our framework demonstrates robust generalization. It substantially mitigates the performance drop observed in global-feature methods (DP, DP3) and single-foundation architectures (GenDP-S, 3DFA). While Pi0.5 (a representative VLA model) exhibits a comparable degradation rate, our absolute success rates remain strictly superior across all settings, validating the efficacy of our decoupled architecture.

This robust generalization stems from the framework's dual capability to handle cross-type interaction diversity and intra-category shape variations. As illustrated in Fig.~\ref{fig:whole_procedure}, HeteroGenManip consistently leads across diverse interaction paradigms (``rigid-articulated'', ``rigid-deformable'', and ``deformable-deformable'') without requiring task-specific architectural adjustments. By employing a multi-foundation routing strategy, the model effectively reconciles the conflicting representational needs of different physical properties. Furthermore, by decoupling initial contact localization from trajectory execution, our model inherently absorbs significant intra-category shape variations for both operated and background objects. For example, in the ``Hang Tops'' task featuring unseen garments and clotheslines, our framework maintains a test success rate of 0.62, far exceeding the best baseline (0.27). This proves that our correspondence-based spatial anchoring provides a highly reliable foundation for complex, different-type object coordination.

\subsection{Ablation Study}
To validate the efficacy of our decoupled architecture, we systematically removed core components and evaluated the performance degradation on the test setting (summarized in Table~\ref{tab:ablation}, and Fig.~\ref{fig:Ablation}). 

\noindent \textbf{Foundation-Correspondence-Guided Grasp Module (w/o CG):} Removing the Foundation-Correspondence-Guided Grasp degrades average performance by 24\%. Without explicit spatial anchoring to resolve initial pose uncertainty, the diffusion policy is forced to learn contact localization and trajectory execution simultaneously. For objects with highly variable grasp states (\emph{e.g.}, the rotating handle in \textit{Store Mug}), this lack of decoupling causes the success rate to severely drop from 0.64 to 0.38.

\noindent \textbf{Position Enhancement Encoder (w/o PE):} Removing the position enhancement encoder causes a 22\% average drop. This confirms that without an explicit spatial coordinate stream, high-dimensional semantic features overshadow fine-grained geometric cues. This impairment is catastrophic for precision-sensitive tasks: in \textit{Beat Block (C)}, spatial striking deviations cause the success rate to plummet by 74\% (from 0.50 to 0.13).

\noindent \textbf{Multi-Foundation-Model (w/o MFM):} Replacing our category-specialized routing with a single uniform model (Uni3D) forces a representational compromise, reducing average success in rigid-deformable tasks. In \textit{Hang Tops}, the rigid-focused Uni3D fails to capture the garment's topological semantics, dropping success from 0.62 to 0.30. Substituting Uni3D with DINOv2 yields even worse results (0.24). This confirms our core insight: cross-type interactions possess asymmetric physical properties that fundamentally require decoupled, multi-foundation representations rather than relying on the capacity of any single model.

\begin{figure}[t] 
  \centering
  \includegraphics[width=\columnwidth]{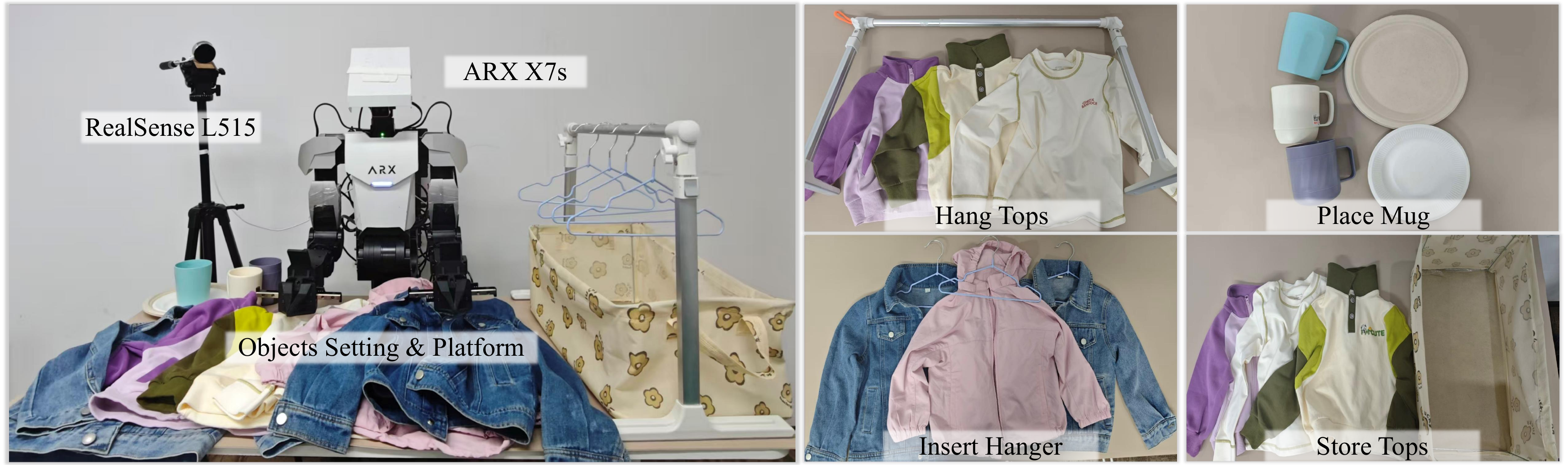} 
  \caption{\textbf{Real-World Setup.} In real-world experiments, we employed the RealSense L515 to capture images and point clouds, while the ARX X7s was utilized for robot manipulation tasks.}
  \label{fig:rw_set} 
  \vspace{-0.5em}
\end{figure}


\begin{table*}[t]
  \centering
  \caption{Real-World Results.}
  \small 
  \renewcommand{\arraystretch}{1.2}
  \begin{tabular*}{\textwidth}{@{\extracolsep{\fill}} l *{10}{c}}
    \toprule
    Method / Task
    & \multicolumn{2}{c}{Hang Tops} 
    & \multicolumn{2}{c}{Insert Hanger} 
    & \multicolumn{2}{c}{Store Tops} 
    & \multicolumn{2}{c}{Place Mug} 
    & \multicolumn{2}{c}{Average} \\ 
    \cmidrule(lr){2-3}  \cmidrule(lr){4-5}  \cmidrule(lr){6-7}  \cmidrule(lr){8-9} \cmidrule(lr){10-11} 
    & Train & Test  & Train & Test & Train & Test & Train & Test & Train & Test \\ 
    \midrule
    \multirow{1}{*}{DP}       
                              & 9/15 & 6/15 & 2/15 & 1/15 & 12/15 & 9/15 & 6/15 & 4/15 & 48.3\% & 33.3\%  \\ 
    \multirow{1}{*}{DP3}     
                              & 8/15 & 6/15 & 3/15 & 2/15 & 12/15 & 10/15 & 7/15 & 6/15 & 50.0\% & 40.0\%  \\

    \multirow{1}{*}{Ours}    
                              & \textbf{12/15} & \textbf{11/15} & \textbf{11/15} & \textbf{9/15} & \textbf{14/15} & \textbf{14/15} & \textbf{13/15} & \textbf{12/15} & \textbf{83.3\%} & \textbf{76.7\%} \\
    \midrule
    \end{tabular*}
    \label{tab:rw}
    \vspace{-1em}
\end{table*}

\begin{figure*}[t!] 
  \centering
  \includegraphics[width=0.95\textwidth]{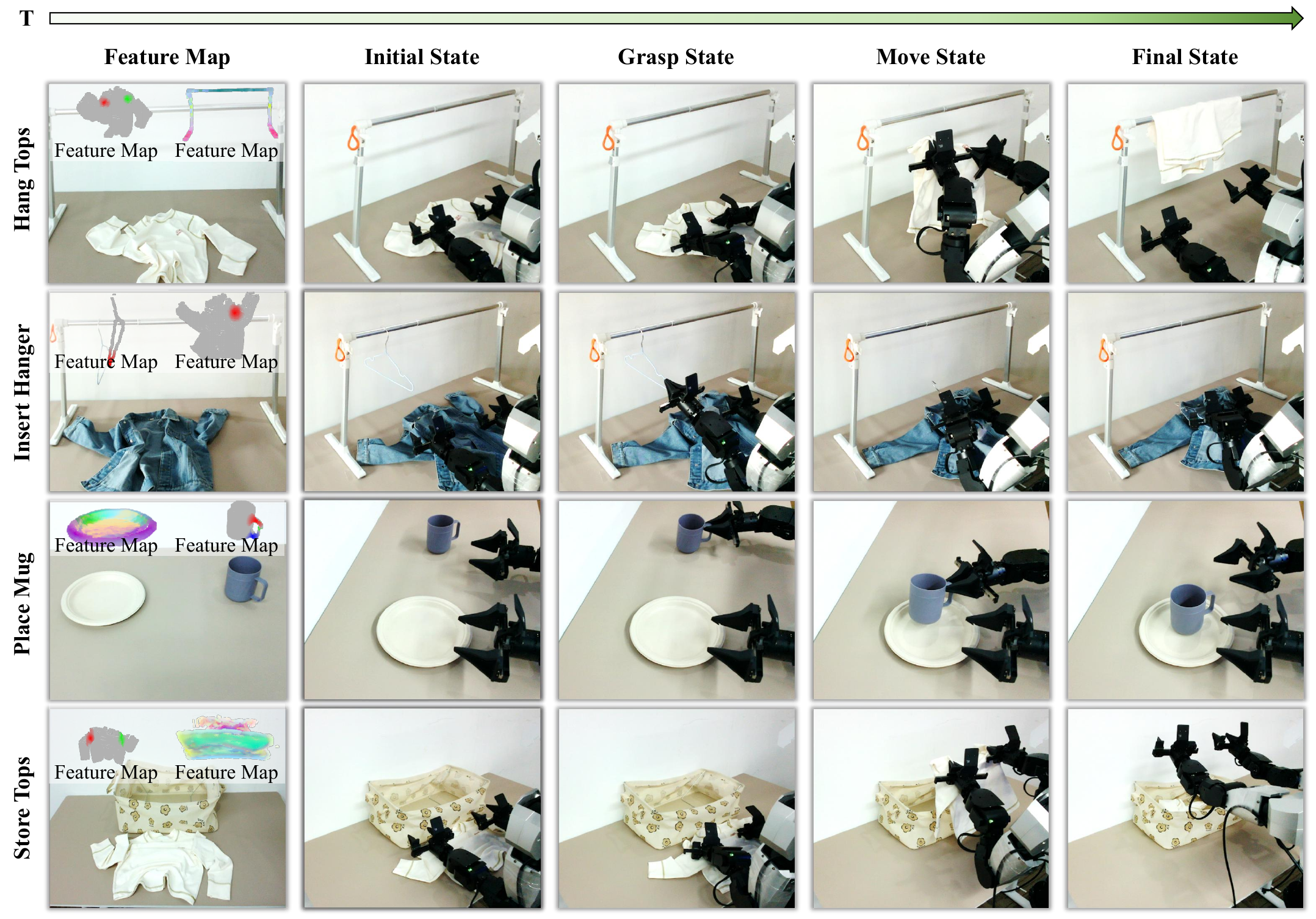} 
  \vspace{-2mm}
  \caption{\textbf{Real-World Execution.} The execution workflow of our four real-world tasks is illustrated in the figure. Specifically, point clouds are collected in the initial state and features are extracted using foundation models. Subsequently, the grasp state and move state are executed sequentially, culminating in the final state to complete the task. The detailed execution description is in Appendix~\ref{appendix:frameworkexec}}.
  
  \label{fig:realworld-procedure} 
  \vspace{-1em}
\end{figure*}

\section{Real-World Experiments}
\label{realworld}
\subsection{Setup}
We select four daily household manipulation tasks with different interaction types: \textit{Hang Tops}, \textit{Place Mug}, \textit{Insert Hanger}, and \textit{Store Tops}, as shown in Fig.~\ref{fig:rw_set}. Detailed task descriptions are provided in Appendix~\ref{appendix:real_tasks}.

For the experimental platform, we employ the ARX-X7s humanoid robot equipped with two 7-DoF arms and two grippers, and use an Intel RealSense L515 depth camera to capture point cloud observations. We collect 25 demonstrations per task via human teleoperation using a Meta Quest 3 device. For dataset arrangement, we use one specific object per task for training data, while two distinct unseen objects are prepared for evaluation. We evaluate each task setting (train or test) independently across 15 trials and report the success rate as the key metric.



\subsection{Results and Analysis}

As shown in table~\ref{tab:rw} and Fig.~\ref{fig:realworld-procedure}, consistent with our simulation findings, our method still outperforms the baseline methods, with 83.3\% on the training setting and 76.7\% on the test setting in term of average success rate.

 This notable performance lead demonstrates that our method possesses a stronger overall capability in real world too. Moreover, the success rate gap between the train and test setting of our method remains relatively narrow: for example, there is no gap at all in the Store Tops task. In contrast, the baselines exhibit larger train-test discrepancies. This result suggests that our method has stronger generalization to unseen test scenarios, effectively avoiding overfitting to the training data. Additional qualitative visualizations are provided in Appendix~\ref{appendix:realworldgene}.

\section{Conclusion}
\label{conclusion}
In this paper, we propose HeteroGenManip to address the generalization challenge in manipulation with heterogeneous object interactions. Our framework comprises Foundation-Correspondence-Guided Grasp (leveraging structural priors for task decomposition and correspondence identification) and Multi-Foundation-Model Diffusion Policy (employing category-specialized models for feature extraction). Together, they eliminate redundant modeling and enable generalization across different-type interactions.
Extensive validation on benchmarks and novel tasks confirms HeteroGenManip's superiority, demonstrating robust generalization to unseen objects and complex interactions in real-world scenarios. This work provides a new paradigm that overcomes limitations of single-model approaches and underutilized structural priors. The limitation of our work is discussed in Appendix~\ref{limit}.

\clearpage
{
\small
\bibliographystyle{plain}
\bibliography{ref}
}
\clearpage
\appendix
\section{Task Description}
\label{appendix:task}

In this appendix, we provide detailed descriptions of all manipulation tasks used in our experiments. We evaluate HeteroGenManip on eight simulation tasks and four real-world tasks, covering diverse interaction types including rigid-rigid, rigid-deformable, rigid-articulated, and deformable-deformable interactions. Each task is designed to test the framework's ability to generalize across different object shapes, poses, and spatial configurations.

\begin{figure*}[ht]
  \centering
  \includegraphics[width=\textwidth]{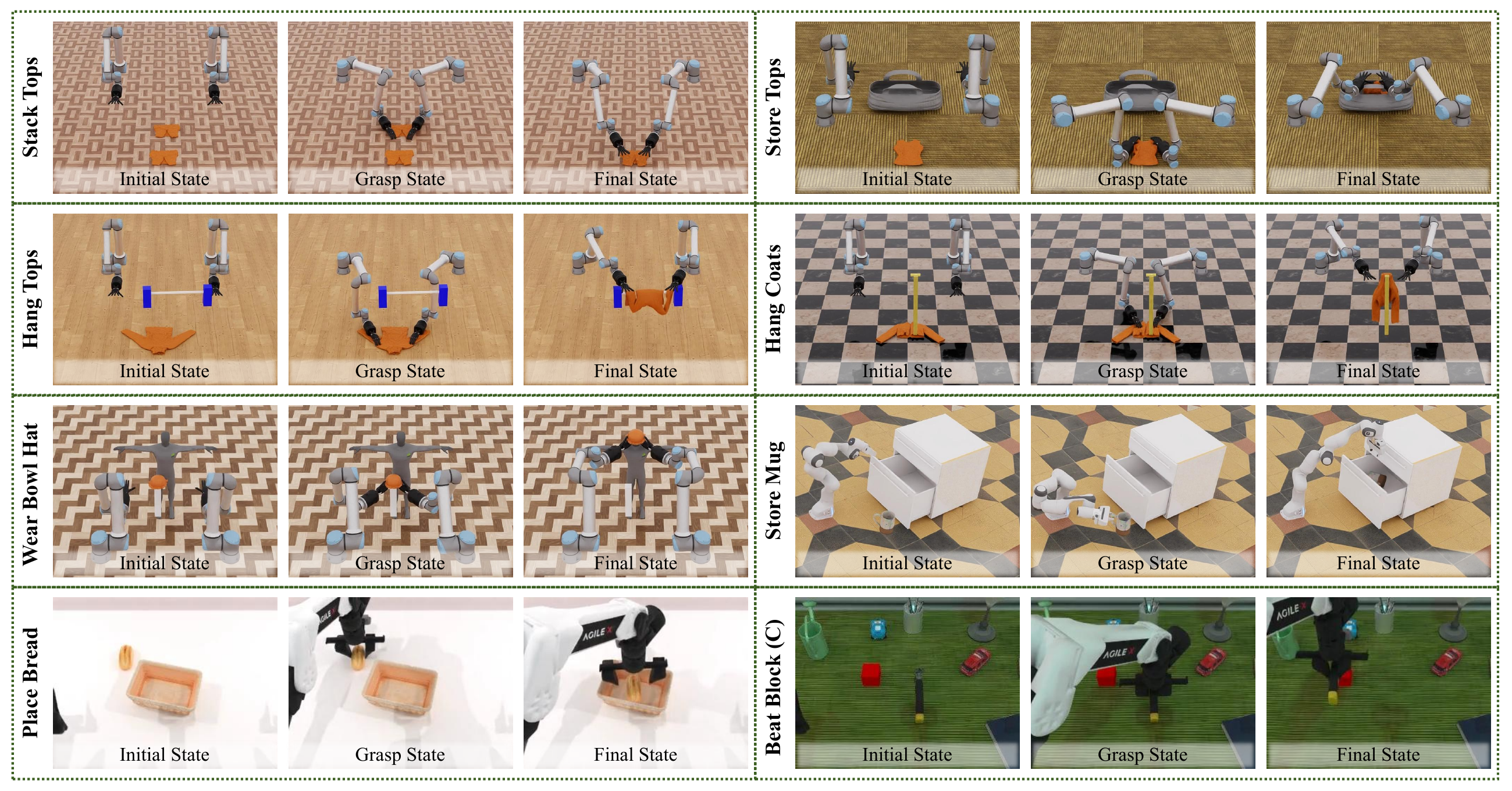}
  \caption{\textbf{Simulation Tasks.} We carefully selected and adapted tasks from DexGarmentLab, RoboTwin and develop novel tasks with different-type object interactions. For each task, we split the data into train and test setting, where the latter imposes higher demands on the generalization ability of the policy.}
  \label{fig:task}
\end{figure*}

\subsection{Simulation Tasks}
\label{appendix:sim_tasks}

As shown in Fig.~\ref{fig:task}, We select and adapt tasks from DexGarmentLab~\cite{wang2025dexgarmentlab} and RoboTwin~\cite{chen2025robotwin}, and develop novel tasks with different-type object interactions. The detailed descriptions are as follows:

\begin{itemize}
    \item \textbf{Stack Tops} (Deformable-Deformable): The robot grasps the lower piece of clothing, moves it to align with another identical top above, and spreads it forward to complete the stacking. The shape of the clothing and their spatial positions are variable.

    \item \textbf{Store Tops} (Rigid-Deformable): The robot grasps a top from its initial flat state, moves it to a position above an upward-opening bag, and slowly places it inside. The shape of the clothing, the shape of the bag, and the spatial positions are variable.

    \item \textbf{Hang Tops} (Rigid-Deformable): The robot grasps a top from its initial flat state and hangs it on a clothesline. The shape of the clothing and clothesline, and the spatial positions are variable.

    \item \textbf{Hang Coats} (Rigid-Deformable): The robot grasps a coat from its initial flat state and hangs it on a vertical hanger. The shape of the clothing and hanger, and the spatial positions are variable.

    \item \textbf{Wear Bowl Hat} (Rigid-Rigid): The robot grasps a bowl hat from its initial placed state, moves it to the position above a person's head, adjusts the posture, and places the hat steadily on the head. The shape of the bowl hat and the spatial positions are variable.

    \item \textbf{Store Mug} (Rigid-Articulated): The robot grasps the handle of a mug (with rotation), then places the mug into a designated storage cabinet. The shape of the mug, the shape of the cabinet, and the spatial positions are variable.

    \item \textbf{Place Bread} (Rigid-Rigid): The robot grasps a loaf of bread and places it into a basket. The shape of the bread, the shape of the basket, and the spatial positions are variable.

    \item \textbf{Beat Block (C)} (Rigid-Rigid, Cluttered): The robot uses a hammer to hit the target block against a cluttered background. The spatial position of the block is variable. The "(C)" denotes a cluttered environment.
\end{itemize}

\subsection{Real-World Tasks}
\label{appendix:real_tasks}

We evaluate our method on four real-world manipulation tasks using the ARX-X7s humanoid robot. These tasks are selected to cover different interaction types and validate the sim-to-real transfer capability of our framework. For each task, we use one object for training and two unseen objects for testing to assess generalization performance. The detailed descriptions are as follows:

\begin{itemize}
    \item \textbf{Hang Tops} (Rigid-Deformable): The gripper grasps the shoulder area of a scattered top, lifts the top to align with the bar of the clothesline, and hangs the top steadily on the clothesline.

    \item \textbf{Insert Hanger} (Rigid-Deformable): One arm retrieves a hanger from the clothesline, while the other arm grasps the collar of the top and lifts it to a suitable height. The two arms then coordinate to insert the hanger through the collar area of the top.

    \item \textbf{Place Mug} (Rigid-Rigid): The gripper grasps the handle of the mug, moves the mug to the position of the plate, and places the mug stably on the surface of the plate.

    \item \textbf{Store Tops} (Rigid-Deformable): The gripper grasps the shoulder area of the top, moves the top to the opening of the storage basket, and places the top neatly inside the basket.
\end{itemize}

\section{Foundation Model Specification}
\label{appendix:foundation_models}
For rigid and articulated objects (characterized by stable geometric structures and consistent affordance patterns), we employ a point-wise variant of the Uni3D~\cite{zhou2023uni3d} model. Departing from the original Uni3D~\cite{zhou2023uni3d} architecture that outputs global object-level features, we modify its final projection layer to generate point-wise features, enabling it to produce a high-dimensional pre-trained feature vector for each point in the input point cloud. This adaptation preserves fine-grained geometric-semantic details at the point level, which are crucial for identifying manipulation-relevant regions (\emph{e.g.}, handles, edges) in rigid and articulated artifacts. 

For deformable objects (exhibiting high dynamicity, shape variability, and context-dependent deformations), we adopt the Garment Affordance Model (GAM)~\cite{wang2025dexgarmentlab}. Unlike rigid \& articulated  object models that focus on absolute feature encoding, GAM is designed to output each point features that prioritize point-corresponding information. This design aligns with the idiosyncrasies of deformable objects, where manipulation relevance is often defined by relational rather than absolute features. During inference, we typically use these features to compute similarity scores relative to pre-defined demonstration points, making this derived similarity a more robust indicator of manipulation suitability.

\section{Training Details}
\label{appendix:training}
Our models were implemented using PyTorch 2.8 and trained on a single NVIDIA GeForce RTX 4090 (24GB) GPU. The detailed hyperparameters used during the training phase are summarized as follows.

\begin{table}[H]  
    \centering
    \caption{Training Hyperparameters for MFMDP}
    \label{tab:training_details}
    \resizebox{\textwidth}{!}{
    \begin{tabular}{lclclc}
        \toprule
        Hyperparameter & Value & Hyperparameter & Value & Hyperparameter & Value \\
        \midrule
        horizon & 8 & n_obs_steps & 3 & n_action_steps & 4 \\
        num_inference_steps & 10 & use_pc_color & False & pca_dim & 5 \\
        optimizer._target_ & torch.optim.AdamW & optimizer.lr & 1.0e-4 & optimizer.betas & [0.95, 0.999] \\
        optimizer.eps & 1.0e-8 & optimizer.weight_decay & 1.0e-6 & training.seed & 42 \\
        training.lr_scheduler & cosine & training.lr_warmup_steps & 500 & training.num_epochs & 3000 \\
        dataloader.batch_size & 128 & dataloader.shuffle & True & dataloader.pin_memory & True \\
        
        \bottomrule
    \end{tabular}
    }
\end{table}

\section{Baseline Training Details}
\label{appendix:baselines}

Regarding the baseline methods, DP, DP3, GenDP-S, and 3DFA were trained using the same hardware configuration (a single RTX 4090 GPU).

\begin{table}[H]  
    \centering
    \caption{Training Hyperparameters for DP, DP3 and GenDP-S}
    \label{tab:DP_DP3_GenDP_details}
    \begin{tabular}{lccc}
        \toprule
        Hyperparameter & DP & DP3 & GenDP-S \\
        \midrule
        horizon & 8 & 8 & 8 \\
        n_obs_steps & 3 & 3 & 3 \\
        n_action_steps & 4 & 4 & 4 \\
        num_inference_steps & 10 & 10 & 10 \\
        use_pc_color & - & False & False \\
        dataloader.batch_size & 32 & 256 & 32 \\
        dataloader.shuffle & True & True & True \\
        dataloader.pin_memory & True & True & True \\
        optimizer._target_ & torch.optim.AdamW & torch.optim.AdamW & torch.optim.AdamW \\
        optimizer.lr & 1.0e-4 & 1.0e-4 & 1.0e-4 \\
        optimizer.betas & [0.95, 0.999] & [0.95, 0.999] & [0.95, 0.999] \\
        optimizer.eps & 1.0e-8 & 1.0e-8 & 1.0e-8 \\
        optimizer.weight_decay & 1.0e-6 & 1.0e-6 & 1.0e-6 \\
        training.seed & 42 & 42 & 42 \\
        training.lr_scheduler & cosine & cosine & cosine \\
        training.lr_warmup_steps & 500 & 500 & 500 \\
        training.num_epochs & 500 & 3000 & 3000 \\
        \bottomrule
    \end{tabular}
\end{table}

\begin{table}[H]  
    \centering
    \caption{Training Hyperparameters for 3DFA}
    \label{tab:3DFA_Pi_details}
    \begin{tabular}{lclc}
        \toprule
        Hyperparameter & Value & Hyperparameter & Value \\
        \midrule
        batch_size & 64 & chunk_size & 1 \\
        lr & 1e-4 & backbone_lr & 1e-4 \\
        train_iters & 600000 & model_type & denoise3d \\
        backbone & clip & embedding_dim & 128 \\
        num_attn_heads & 8 & num_shared_attn_layers & 4 \\
        num_vis_instr_attn_layers & 3 & num_history & 1 \\
        denoise_timesteps & 10 & denoise_model & rectified_flow \\
        \bottomrule
    \end{tabular}
\end{table}
Unlike the other baselines, the Pi0.5 baseline, we initialized the model with official pre-trained weights and fine-tuned it via LoRA using 8 NVIDIA A800 GPUs. Notably, for dexterous-hand tasks, we compress the hand state into a one-dimensional variable, where 0 and 1 respectively denote the open and closed states, and intermediate discrete values correspond to linearly interpolated positions between them.
\begin{table}[H]  
    \centering
    \caption{Training Hyperparameters for Pi0.5}
    \label{tab:3DFA_Pi_details}
    \begin{tabular}{lclc}
        \toprule
        Hyperparameter & Value & Hyperparameter & Value \\
        \midrule
        batch_size & 1320 & num_train_steps & 800 \\
        warmup_steps & 90 & decay_steps & 710\\
        decay_lr & 2e-05 & peak_lr & 0.0008 \\
        optimizer.betas & [0.9, 0.98] & ema_decay & 0.995\\
        model.paligemma_variant & gemma_2b_lora & model.action_expert_variant & gemma_300m_lora \\
        model.action_horizon & 12 & model.max_token_len & 200 \\
        \bottomrule
    \end{tabular}
\end{table}

\section{Framework Execution Procedure}
\label{appendix:frameworkexec}

Our proposed framework operates in a task-conditioned manner. Given a task description, we embed it into a specific prompt and feed it, along with the image of the task's initial state, into a Vision-Language Model (VLM), specifically Qwen2.5-VL-7B\cite{bai2025qwen25vltechnicalreport}. The VLM parses the task content and outputs the names and categories of both the operated object and the background object. These categories are selected from three predefined types: deformable, rigid, and articulated. The assigned category dictates which foundation model should be applied to each respective object. Subsequently, the extracted object names serve as text prompts, which are fed into the SAM 3 model\cite{carion2025sam3segmentconcepts} alongside the same initial state image to generate pixel-level masks for each object. By combining these 2D masks with the global point cloud, we extract the segmented 3D point cloud for each object. Finally, these segmented point clouds, along with the global point cloud, are utilized to execute our downstream pipeline.

\section{Real-World Generalization Visualization}
\label{appendix:realworldgene}
\begin{figure*}[h] 
  \centering
  \includegraphics[width=\textwidth]{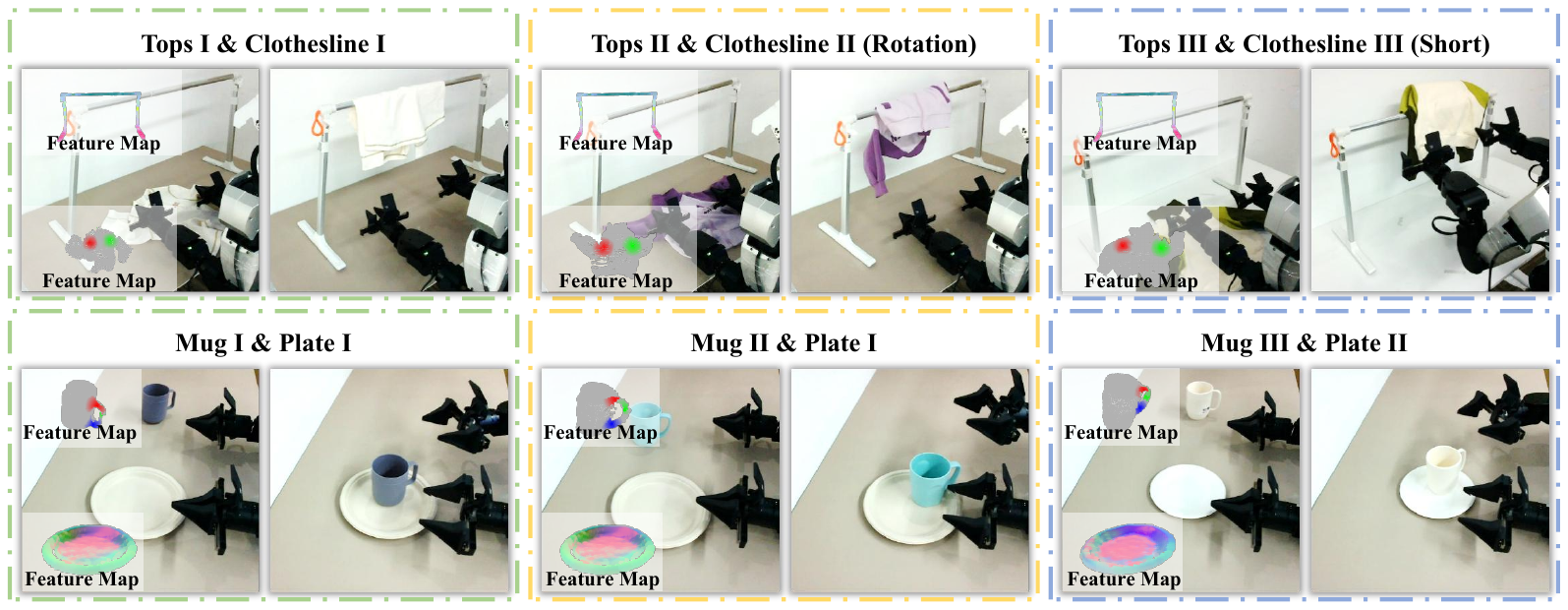} 
  \vspace{-5mm}
  \caption{\textbf{Generalization in Real-World Manipulation.} This figure qualitatively demonstrates the generalization performance of our method in real-world manipulation tasks. Taking two tasks (Hang Tops and Place Mug) as examples, it presents our tests on diverse object configurations, alongside the correspondence and semantic feature maps.}
  \vspace{-5mm}
  \label{fig:realworld-show} 
\end{figure*}

To further validate the generalization capability of our method beyond simulation, we conduct real-world manipulation experiments on diverse object configurations. Fig.~\ref{fig:realworld-show} presents qualitative visualizations of two representative tasks: \textit{Hang Tops} and \textit{Place Mug}, demonstrating our method's ability to handle variations in object appearance, pose, and spatial arrangement.

Across both tasks, the visualizations demonstrate three critical aspects of our method's generalization: (1) \textit{Correspondence consistency}: The correspondence maps maintain stable anchor points across object variations, confirming that our foundation-model-driven correspondence provides robust spatial grounding. (2) \textit{Semantic discrimination}: The feature maps show distinct activation patterns for different object categories (deformable vs. rigid), validating our multi-foundation model routing strategy. (3) \textit{Cross-shape transfer}: The method successfully transfers from training shapes to novel test shapes without fine-tuning, demonstrating the effectiveness of our decoupled architecture in leveraging pre-trained foundation models for generalization.

These real-world results complement our simulation experiments and provide empirical evidence that our method can translate effectively to real robot manipulation scenarios.

\section{Limitation}
\label{limit}
Although our manipulation framework has demonstrated excellent performance across various types of tasks, it still has several limitations. First, the input point clouds we use are all single-view point clouds, so the extraction of object semantic features using these point clouds is highly dependent on the capability of foundation models and camera perspectives. Second, our framework requires object point clouds as input, which necessitates segmenting the target with a 2D vision model first and then projecting it into point cloud; however, the segmentation results and depth map quality for some objects (such as thin-edged hangers and clotheslines) are poor, which severely limits the real-world performance of our framework.

\end{document}